\documentclass[11pt]{article}
\usepackage{amsmath}
\usepackage{amsfonts}
\usepackage{varwidth}
\usepackage{epsf}
\usepackage{algorithm}
\usepackage{algorithmic}
\usepackage{xspace}
\usepackage{booktabs}
\usepackage{eucal,bm}
\usepackage{setspace}
\input{bookstuff.sty}
\usepackage{hyperref}
\hypersetup{
   bookmarks=true,         % show bookmarks bar?
   pdfauthor={},     % author
   colorlinks,%
   citecolor=black,%
   filecolor=black,%
   linkcolor=black,%
   urlcolor=black
}
\usepackage[all]{hypcap}

\def\l{\lambda}
\def\g{\gamma}
\def\a{\alpha}
\def\d{\delta}

\renewcommand{\e}{{\bm e}}
\def\ph{\bm\phi}
\def\th{\bm\theta}
\def\la{($\l$)\xspace}
\def\gets{\leftarrow}

\title{True Online Emphatic TD\la: \\
Quick Reference and Implementation Guide}
\author{Richard S. Sutton}
\date{Revised \today}

\begin{document}
\maketitle

This document is a guide to the implementation of \emph{true online emphatic TD\la}, a model-free temporal-difference algorithm for learning to make long-term predictions which combines the emphasis idea (Sutton, Mahmood \& White 2015) and the true-online idea (van Seijen \& Sutton 2014). The setting used here includes linear function approximation, the possibility of off-policy training, and all the generality of general value functions (Maei \& Sutton 2010), as well as the emphasis algorithm's notion of ``interest". Conventional TD\la is of course the core model-free algorithm for learning value functions in reinforcement learning (Sutton 1988, Sutton \& Barto 1998). The emphasis idea is to dynamically rescale the updates made by temporal-difference algorithms such that convergence is ensured under off-policy training (Yu 2015) and such that asymptotic accuracy of the approximation is improved. The true-online idea extends TD\la to make it more data efficient and less sensitive to step-size settings, at minimal computational expense (van Seijen, Mahmood, Pilarski \& Sutton 2015). The way that these ideas have been combined to produce true online emphatic TD\la was modelled after how van Hasselt, Mahmood, and Sutton (2014) combined the true-online idea and the gradient-TD idea (Maei 2011, Sutton et al.~2009) to produce true online GTD\la.

\section{Setting and requirements}

We consider the setting of general value functions, or GVFs (Maei \& Sutton 2010, Sutton et al.~2011, White 2015, Sutton, Mahmood \& White 2015). Here we present these ideas without assuming access to an underlying state (as in Modayil, White \& Sutton 2014).

The algorithm is meant to be called at regular intervals with data from a time series, from which it learns to make a prediction. The time series includes a feature vector $\ph_t\in\Re^n$ and a cumulant signal $R_t\in\Re$. 
\[ \ph_0, R_1, \ph_1, R_2, \ph_2, R_3, \ph_4, \ldots \]
The prediction at each time is linear in the feature vector. That is, the prediction at time $t\ge0$ is of the form 
\[
\ph_t\tr\th_t = \sum_{i=1}^n \phi_t(i) \theta_t(i), 
\]
where $\th_t\in\Re^n$ is a learned weight vector at time $t$, and $\phi_t(i)$ and $\theta_t(i)$ are of course the $i$th components of the corresponding vectors. The learning process results in the prediction at each time $t$ coming to approximate the outcome, or target, that would follow it:
%$$\ph_t\tr\th_t \approx \CE{G_t}{A_{t:\infty}\sim\pi} = \CE{\sum_{k=t+1}^\infty \! R_k \prod_{j=t+1}^{k-1} \g(S_j)}{A_{t:\infty}\sim\pi},$$
\[
\ph_t\tr\th_t \approx \sum_{k=t+1}^\infty \! R_k \prod_{j=t+1}^{k-1} \g_j
\]
if actions were selected according to policy $\pi$, and where $\g_t\in[0,1]$ is a sequence of discount factors. We see from this equation why the signal $R_t$ is termed the ``cumulant"; all of its values are added up, or \emph{accumulated}, within the temporal envelope specified by the $\g_j$. In the special case in which the cumulant is a reward and the $\g_j$ are constant then the GVF reduces to a conventional value function from reinforcement learning.

To make the GVF problem well defined, the user must provide $\pi$ and the $\g_j$. The policy $\pi$ is not provided directly, but in the form of a sequence of importance sampling ratios
\[
\rho_t = \frac{\pi(A_t|S_t)}{\mu(A_t|S_t)},
\]
where $S_t$ and $A_t$ are the state and action actually taken at time $t$, and $\pi(A_t|S_t)$ and $\mu(A_t|S_t)$ are the probabilities of $A_t$ in $S_t$ under policies $\pi$ and $\mu$ respectively. The policy $\pi$ is called the \emph{target policy}, because it is under it that we are trying to predict the outcome, as stated above, and $\mu$ is called the \emph{behavior policy}, because it is it that actually generates the behavior and the time series.
Because only the ratio of the two probabilities is required, there is often no need to work directly with states or action probabilities. For example, in the on-policy case the target and behavior policies are the same, and the ratio is always one. The discount factors are often taken to be constant, but are allowed to depend arbitrarily on the time series, as long as $\prod_{j=t+1}^\infty \g_j = 0$ for all $t$.

In some publications concerning general value functions there is also specified a fourth sequence pertaining to the prediction problem---the ``terminal pseudo reward" $Z_t$---to specify a final signal to be added in with the cumulants at termination. More recently its has been recognized that this functionality can be included with just the cumulant $R_t$ by appropriately setting the discount sequence $\g_t$ (see Modayil, White \& Sutton 2014). For example, if one wanted a terminal pseudo reward of $Z_t$ only upon termination, then one would use a cumulant of $R_t = (1-\g_t)Z_t$. 

In addition to the time series of the feature vectors and cumulant signals, the user must provide three sequences characterizing the nature of the approximation to be found by the algorithm:
\begin{itemize}
\item $I_t\ge0$; the \emph{interest sequence} specifies the interest in or importance of accurately predicting at time $t\ge0$. For example, in episodic problems one may care only about the value of the first state of the episode; this is specified by setting $I_t=1$ for the first state of each episode and $I_t=0$ at all other times. (Or, as suggested by the work of Thomas (2014), one may want to use $I_t=\g^t$, where $t$ here is the time since the beginning of the episode.)
In a discounted continuing task, on the other hand, one often cares about all the states equally, which is specified by setting $I_t=1$ for all $t$. In general, if one has any reason to be more concerned with the approximation being more accurate at some times than others, this can be expressed through the interest sequence.
\item $\l_t\in[0,1]$; the \emph{bootstrapping} sequence specifies the degree of bootstrapping at each time.
\item $\a_t\ge 0$; the step-size sequence specifies the size of the step at each time. One common choice is a constant step-size parameter, e.g., $\a_t = 0.1/\max_t \ph_t\tr\ph_t$. Another common choice is a step-size parameter that decreases to zero slowly over time. More sophisticated step-size adaptation methods could also be used to determine the step-size sequence (e.g., Mahmood et al.~2012, Dabney \& Barto 2012, Reidmiller \& Braun 1993)
\end{itemize}

\section{Algorithm Specification}

Internal to the learning algorithm are the learned weight vector, $\th_t\in\Re^n$, and an auxiliary shorter-term-memory vector $\e_t\in{\Re}^n$ with $\e_t\ge\bm 0$. In addition, there are the scalars $M_t\ge 0$ and $F_t\ge 0$. The emphasis $M_t$ and the TD error $\d_t$ are purely temporary variables. The true online emphatic TD\la algorithm is fully specified by the following equations:
\begin{align}
 \d_t &= R_{t+1} + \g_{t+1}\th_t\tr\ph_{t+1} - \th_t\tr\ph_t\\
 F_t &= \rho_{t-1}\g_t F_{t-1} + I_t, \hspace*{150pt}\text{with~}F_{-1}=0 \\
 M_t &=  \l_t \, I_t + (1-\l_t) F_t \\
 \e_t &= \rho_t\g_t\l_t\e_{t-1} + \rho_t\a_t M_t(1-\rho_t\g_t\l_t\ph_t\tr\e_{t-1})\ph_t\text{~~~~~~~~with~} \e_{-1}=0\\
 \th_{t+1} &= \th_t + \d_t\e_t + (\e_t-\a_t M_t\rho_t\ph_t)(\th_t-\th_{t-1})\tr\ph_t
\end{align}

%\vspace{-.2in}
\section{Pseudocode}

%\vspace{-.1in}
The following pseudocode characterizes the algorithm and its efficient implementation in C++. First the {\tt init} function should be called with argument $n$ (the number of components of $\th$ and $\ph$):

\def\trc{\!\tr\!\!}
\def\u#1{{\underbar{$#1$}}}
\noindent\fbox{
\begin{varwidth}{\dimexpr\linewidth-2\fboxsep-2\fboxrule\relax}
\begin{tabbing}
~~~\=\kill
{\tt init($n$):}\\
\>store $n$ \\
\>$\e \gets \bm 0$ \\
\>$\th \gets \bm 0$ ~~~~~(or arbitrary)\\
\>$F \gets D \gets \g \gets 0$
%\>$\u{\th\trc\ph} \gets \th\tr\ph$ \\
\end{tabbing}
\end{varwidth}
}

\noindent
%The underlined groups of symbols are simply scalar variables (whose content is suggested by the symbols), that are saved from one time step to the next.
On each step,  $t=0, 1, 2, \ldots$, the {\tt learn} 
function is called with arguments $\a_t, I_{t}, \l_{t}, \ph_t, \rho_{t}$, $R_{t+1}, \ph_{t+1}, \g_{t+1}$:

\noindent\fbox{
\begin{varwidth}{\dimexpr\linewidth-2\fboxsep-2\fboxrule\relax}
\begin{tabbing}
~~~\=~~~~~~~~~~~~~~~~~~~~~~~~~~~~~~~~~~~~~~~~~\=\kill
{\tt learn($\a, I, \l, \ph, \rho, R, \ph', \g'$):} \>\>; $\a$ thru $\rho$ are at $t$, the rest are at $t+1$\\[3pt]
\> $\d \gets R + \g'\th\tr\ph' - \th\tr\ph$ \> ; or, do all 3 inner products in a single loop\\[3pt]
\> $F \gets F + I$ \>; $F$ was $\rho_{t-1}\g_t F_{t-1}$; now it is $F_t$ \\[3pt]
\> $M \gets \l I + (1-\l)F$ \\[3pt]
\> $S \gets \rho\,\a M(1-\rho\g\l\ph\tr\e)$ 
 \>; scalar $S$ saves computation \\[3pt]
\> $\e \gets \rho\,\g\l\e + S\ph$ \> ; this + next 3 lines can be done in a single loop\\[3pt]
%\> $\e \gets \rho\left[\l\e + \a M (1-\rho\l\, \u{\g\e\trc\ph}) \ph\right]$ 
% \>; the 2nd $\e$ is $\g_t\e_{t-1}$; $\u{\g\e\trc\ph}$ is $\g_{t}\e_{t-1}\tr\ph_t$ \\[2pt]
\> $\Delta \gets \d\e + D(\e-\rho\,\a M\ph)$ 
 \>; $D$ here is $(\th_t-\th_{t-1})\tr\ph_t$\\[3pt]
\> $\th \gets \th + \Delta$ \\[3pt]
%\> $\e \gets \g\e$ \>; $\e$ is now $\g_{t+1}\e_t$\\[1pt]
%\> $\u{\g\e\trc\ph} \gets \e\tr\ph'$ \>; $\u{\g\e\trc\ph}$ is now $\g_{t+1}\e_t\tr\ph_{t+1}$\\[1pt]
\> $D \gets \Delta\tr\ph'$ 
% \>; $\u{\Delta\trc\ph}$ here is $(\th_{t+1}-\th_{t})\tr\ph_{t+1}$ 
\\[3pt]
\> $F \gets \rho\,\g' F$ %\>; $F$ is now $\rho_t\g_{t+1}F_t$ 
\\[3pt]
\> $\g \gets \g'$
\end{tabbing}
\end{varwidth}
}
\goodbreak

%\vspace{100.00cm}
%\noindent\fbox{
%\begin{varwidth}{\dimexpr\linewidth-2\fboxsep-2\fboxrule\relax}
%\begin{tabbing}
%Initialize $\th$ arbitrarily and $\w = 0$ \\
%Repeat \=(for each episode):\\
%\> Initialize $\e = 0$\\
%\> $S \gets$ initial state of episode \\
%\> Repeat \=(for each step of episode):\\
%\>\>$A \gets$ action selected by policy $b$ in state $S$ \\
%%\>\> $\ph_{s,a} \gets$ set of features present in $s,a$\\ 
%\>\>Take action $A$, observe next state, $S^\prime$\\
%\>\> $\bar\ph \gets 0$\\
%\>\> For \=all $a \in \A(s)$:\\
%\>\>\> $\bar\ph \gets \bar\ph + \pi(S^\prime,a)\ph(S^\prime,a)$\\ 
%\>\>$\rho = \frac{\pi(S, A)}{b(S, A)}$\\
%\>\>GQlearn($\ph(S,A), \bar\ph, \l(S^\prime), \g(S'), r(S, A, S^\prime), \rho, I(S)$)\\
%\>\>$S \gets S^\prime$\\
%\> until $S^\prime$ is terminal
%\end{tabbing}
%\end{varwidth}
%}

\noindent
Finally, to obtain a prediction based on the learned weights, pass a feature vector to the {\tt predict} function:

\noindent\fbox{
\begin{varwidth}{\dimexpr\linewidth-2\fboxsep-2\fboxrule\relax}
\begin{tabbing}
~~~\=\kill
{\tt predict($\ph$):}\\
\>return $\th\tr\ph$
\end{tabbing}
\end{varwidth}
}

If the task is episodic in the classical sense, then the terminal state should be represented as a special additional state at which $\g=0$, $\ph=\bm 0$, and with outgoing transitions to the distribution of start states. As far as {\tt learn} is concerned, there is still just a single sequence.

%\section{Action-value form}
%
%So far we have treated the algorithm for learning a state-value function.
%To learn an action-value function, no changes are needed in either the algorithm or its implementations. One merely re-interprets the feature vector at time $t$ as corresponding to the state--action pair at $t$ rather than solely to the state. But then there may be a few other small changes to be worked out...
%

\section{Code}

Implementations that closely follow the pseudocode are provided for various programming languages in separate files. Where we have seen it as convenient and non-obfuscating, the implementations are in an object-oriented style in which one creates an instance of the algorithm that contains all of its internal variables.

\section*{Acknowledgements}

The author gratefully acknowledges the assistance of Hado van Hasselt, Harm van Seijen, and A. Rupam Mahmood in preparing this guide.

\section*{References}
\parindent=0pt
\def\hangin{\hangindent=0.15in}
\parskip=6pt

\hangin
Dabney, W., Barto, A. G. (2012). Adaptive step-size for online temporal difference learning. In \emph{Proceedings of the Conference of the Association for the Advancement of Artificial Intelligence} (AAAI).

\hangin
Maei, H.~R. (2011).
\emph{Gradient Temporal-Difference Learning Algorithms}. 
PhD thesis, University of Alberta. 

\hangin
Maei, H.~R., Sutton, R.~S. (2010). 
GQ($\l$): A general gradient algorithm for temporal-difference prediction learning with eligibility traces. 
In  \emph{Proceedings of the Third Conference on Artificial General Intelligence}, pp.~91--96. Atlantis Press.

\hangin
Mahmood, A. R., Sutton, R. S., Degris, T., Pilarski, P. M. (2012). Tuning-free step-size adaptation. In \emph{Proceedings of the IEEE International Conference on Acoustics, Speech and Signal Processing} (ICASSP), pp. 2121-2124. IEEE Press.

\hangin
Modayil, J., White, A., Sutton, R.~S. (2014). Multi-timescale nexting in a reinforcement learning robot. \emph{Adaptive Behavior 22}(2):146--160.

\hangin
Riedmiller, M., Braun, H. (1993). A direct adaptive method for faster backpropagation learning: The RPROP algorithm. In \emph{Proceedings of the IEEE International Conference on Neural Networks} (pp. 586-591). IEEE Press.

%\hangin
%Singh, S.~P., Sutton, R.~S. (1996).
%\newblock Reinforcement learning with replacing eligibility traces.
%\newblock {\em Machine Learning}, 22:123--158.

\hangin
Sutton, R. S. (1988). Learning to predict by the methods of temporal differences. \emph{Machine Learning 3}:9--44.

\hangin
Sutton, R. S., Barto, A. G. (1998). \emph{Reinforcement Learning: An Introduction}. MIT Press.

\hangin
Sutton, R.~S., Maei, H.~R., Precup, D.,  Bhatnagar, S., Silver, D., Szepesv{\'a}ri, {Cs}.,  Wiewiora, E. (2009).
Fast gradient-descent methods for temporal-difference learning with linear function approximation.
In \emph{Proceedings of the 26th International Conference on Machine Learning}, pp. 993--1000, ACM.

\hangin
Sutton, R. S., Mahmood, A. R., White, M. (2015).
An emphatic approach to the problem of off-policy temporal-difference learning.
ArXiv:1503.04269.

\hangin
Sutton, R.~S., Modayil, J., Delp, M., Degris, T., Pilarski, P.~M., White, A., Precup, D. (2011). Horde: A scalable real-time architecture for learning knowledge from unsupervised sensorimotor interaction. In \emph{Proceedings of the 10th International Conference on Autonomous Agents and Multiagent Systems}, pp. 761--768.

\newpage
\hangin
Thomas, P. (2014).
Bias in natural actor--critic algorithms.
In \emph{Proceedings of the 31st International Conference on Machine Learning}.
JMLR W\&CP 32(1):441--448.

\hangin
van Hasselt, H., Mahmood, A. R., Sutton, R. S. (2014). Off-policy TD\la with a true online equivalence. In \emph{Proceedings of the 30th Conference on Uncertainty in Artificial Intelligence}, Quebec City, Canada.

\hangin
van Seijen, H., Sutton, R. S. (2014). True online TD\la. In \emph{Proceedings of the 31st International Conference on Machine Learning}. Beijing, China. JMLR: W\&CP volume 32.

\hangin
van Seijen, H., Mahmood, A. R., Pilarski, P. M., Sutton, R. S. (2015).
An empirical evaluation of true online TD\la.
In \emph{Proceedings of the 2015 European Workshop on Reinforcement Learning}.

\hangin
Yu, H. (2015).
On convergence of emphatic temporal-difference learning.
In \emph{Proceedings of the Conference on Computational Learning Theory}.

\hangin
White, A. (2015).
\emph{Developing a Predictive Approach to Knowledge}.
Phd thesis, University of Alberta.

\end{document}